\documentstyle[11pt,epsfig]{article}
\newcommand{\be}{\begin{equation}}
\newcommand{\ee}{\end{equation}} 
\newcommand{\eei}{\end{equation}\indent\indent}
\newcommand{\bc}{\begin{center}}

\newcommand{\ec}{\end{center}}
\newcommand{\ber}{\begin{eqnarray}}
\newcommand{\ear}{\end{eqnarray}}
\newcommand{\ba}{\begin{array}}
\newcommand{\ea}{\end{array}}

\def\case#1/#2{\textstyle\frac{#1}{#2} }
\begin{document}
\title{Ultrametric Distance in Syntax}
\author{Mark D. Roberts, \\
117 Queens' Road, Wimbledon, London SW19 8NS\\
Email mdr@ic.ac.uk,  http://cosmology.mth.uct.ac.za/$\sim$ roberts}
\maketitle
\bc Eprint:  http://xxx.lanl.gov/abs/cs.CL/9810012
             http://cogprints.soton.ac.uk/abs/cog00000225 \ec
    Comments:  28 pages  55508 bytes,  
Sixteen EPS diagrams,  39 references, \\
some small changes from the previous version,  matrices reset,\\
background to this work can be found at:\\
http://cosmology.mth.uct.ac.za/~roberts//pastresearch/ultrametric.html
\bc Keyword Index:
Phrase Trees,~~Ultrametricity,~~Syntax,
~~Definitions of Government,~~Features, ~~\={X} Structure.\ec
\bc ACM Classification:  
http://www.acm.org/class/1998/overview.html  
I.2.7;  J.4;  I.2.6  \ec
\bc Mathematical Review Classification: 
http://www.ams.org/msc/  
03D65, 68S05, 92K20, 03D55, 92B10.\ec
\begin{abstract}
Phrase structure trees have a hierarchical structure. 
In many subjects,  most notably in {\bf taxonomy} such tree structures 
have been studied using ultrametrics.   
Here syntactical hierarchical phrase trees are subject 
to a similar analysis,  which is much simpler as the branching structure is 
more readily discernible and switched.  The occurrence of hierarchical 
structure elsewhere in linguistics is mentioned.   
The phrase tree can be represented 
by a matrix and the elements of the matrix can be represented by triangles.
The height at which branching occurs is not prescribed in previous syntactic
models,  but it is by using the ultrametric matrix.   
In other words the ultrametric approach gives a complete description of
phrase trees,  unlike previous approaches.
The ambiguity of which branching height to choose,
is resolved by postulating that branching occurs 
at the lowest height available.   An ultrametric produces a measure of the
complexity of sentences:  presumably the complexity of sentences increases 
as a language is acquired so that this can be tested.   All ultrametric 
triangles are equilateral or isosceles,  here it is shown that \={X}
structure implies that there are no equilateral triangles.   Restricting
attention to simple syntax a minimum ultrametric distance between lexical 
categories is calculated.   This ultrametric distance is shown to be 
different than the matrix obtained from features.   It is shown that the 
definition of {\sc c-command} can be replaced by an equivalent ultrametric 
definition.   The new definition invokes a minimum distance between nodes and 
this is more aesthetically satisfying than previous varieties of definitions.
From the new definition of {\sc c-command} follows a new definition of
{\sc government}.
\end{abstract}
{\small\tableofcontents}
\section{Introduction.}\label{sec:intro}
\subsection{Ultrametrics.}
Ultrametrics are used to model any system that can be represented
by a bifurcating hierarchical tree.
The is a relationship between trees annd ultrametrics is as follows.
An $N$-leaf edge(node)-weighted tree corresponds to an $N\times N$ 
square matrix $M$ in which $M_{ij}=$ the sum of the weights of the edges(nodes)
in the shortest path between $i$ and $j$.
When the weights are non-negative,  $M$ ia a measure in the usual sense when
\ber
&&\forall x,y,z~~~M_{xy}=0~~~ {\rm if}~~~ x=y\\
&&M_{xy}>0~~~ {\rm for}~~~ x\ne y\\
&&M_{xy}=M_{yx}\\
&&M_{xy}\le M_{xz}+M{zy},\label{triaineq}\\
\nonumber
\ear
if the traingle inequality \ref{triaineq} is replaced by 
\be
M_{xy}\le {\rm max}\{M_{xz},M_{zy}\}.
\label{ultramax}
\ee
then $M$ is an ultrametric.
To briefly go through some areas where ultrametrics 
have been applied.
Perhaps the most important application is to taxonomy,
Jardine and Sibson (1971) Ch.7 \cite{bi:JS}, 
and Sneath and Sokal (1973) \cite{bi:SS}.
Here the end of a branch of the tree represents a species and the ultrametric
distance between them shows how closely the species are related.
The actual technique is called the hierarchical cluster method,  
the method classifies species and also shows how closely species are related.
This technique has also been used in semantics,
Shepard and Arabie (1979) \cite{bi:SA}.
The technique can become quite complex because they 
involve statistical analysis with continuous variates.
Ultrametrics have been applied frequently in the theory of spin glass,
Weissman \cite{bi:weissman}.
Ultrametrics have been used for description of 
slowly driven dissipative systems,
which exhibit avalanche-like behaviour,  these include earthquakes,
extinction events in biological evolution,  and landscape formation,
Boettcher and Paiginski (1997) \cite{bi:BP},
also ultrametrics can describe systems with fast relaxation,
Vlad (1994) \cite{bi:vlad}.
Ultrametrics are frequently used in the theory of neural nets,
Parga and Virasoro \cite{bi:PV}.
The dynamics of random walks on ultrametric spaces have been studied,
Ogielchi and Stein (1985) \cite{bi:OS}.
Ultrametrics have been applied to the thermodynamics of macromolecules 
such as RNA,  Higgs (1996) \cite{bi:higgs}.
Bounds on the size of ultrametric structure have been discussed by
Baldi and Baun (1986) \cite{bi:BB}.
From a more theoretical angle,  a category theory approach 
has been elucidated by Rutten (1996) \cite{bi:rutten},  
and a model theoretic approach to them given Delon (1984) \cite{bi:delon}.
The relationship between ultrametric distance and hierarchy is further 
discussed in Gu\'{e}noche (1997) \cite{bi:guenoche}.
Construction of optimal ultrametric trees is discussed by
Young and DeSarbo (1995) \cite{bi:YDS}.
Ultrametrics are related to p-adelic quantities, 
Karwowski and Mendes (1994) \cite{bi:KM}.
P-adelic quantities are used in string theory, 
the way that ultrametrics enters here is explained in \S10\&\S13.4 of
Bekke and Freund (1993) \cite{bi:BF}.
There does not seem to be any straightforward connection of any of the above
to the optimization techniques of
Prince and Smolensky (1997) \cite{bi:smolensky}.
As well as ultrametric trees,  
there are also {\sl decision trees} Hammer (1998) \cite{bi:hammer},
and the connection between them is still not known.
Some of the above ultrametric applications have been reviewed by
Rammal {\it et al} (1986) \cite{bi:rammal}
\subsection{Syntactic Phase Trees.}
For the analysis of syntactic phrase 
trees the necessary technique is quite simple and 
is illustrated by the examples in section
\ref{sec:sec2}.   The examples here mainly follow the examples in 
Lockward (1972) \cite{bi:lockward}, Kayne (1981) \cite{bi:kayne},
McCloskey (1988) \cite{bi:mccloskey},  
and especially Haegeman (1994) \cite{bi:haegeman}.
There are at least {\bf five} reasons for introducing an ultrametric 
description of syntax.\\
The {\sf first} is to completely specify tree 
(also called {\sl dendrogram}) structure.
Consider the following example illustrated by {\bf Figure 1}:
\begin{figure}
{\epsfig{figure=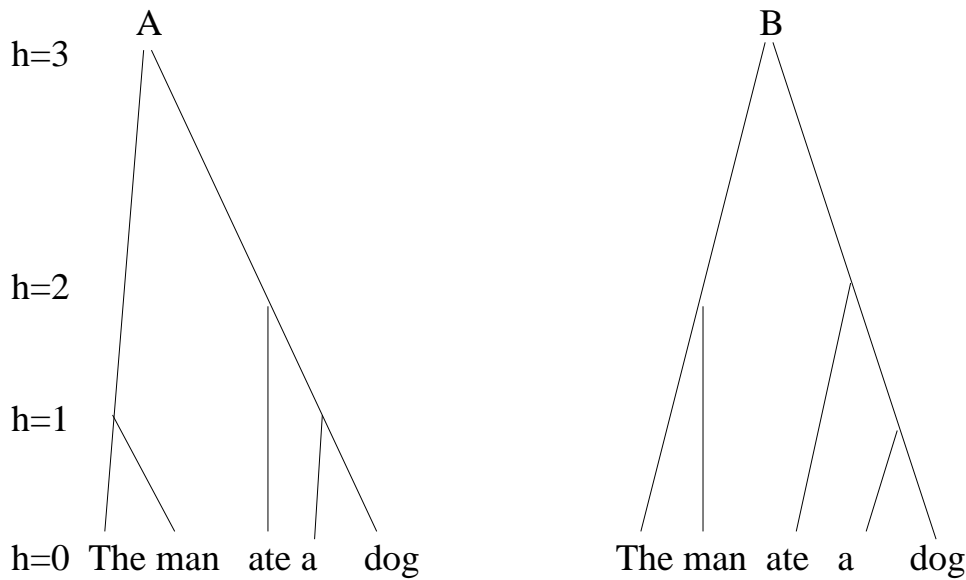,height=3in}}
\caption{Different syntactic descriptions of ``the man ate a dog''}
\label{DiagramOne}
\end{figure}
For current syntactic models the two trees usually are equivalent
(perhaps not always McCloskey (1988) \cite{bi:mccloskey} footnote 6):  
however consider the ultrametric distance between 'the' and 'man',
\be
A(the,man)=1,~~~~~~
B(the,man)=2,
\label{eq:udtheman}
\ee
this difference does not occur in current syntactic models,
and a purpose of an ultrametric model is to disambiguate this.\\
The {\sf second} is it gives a measure of the complexity of a sentence:
the greater the ultrametric distance required the more complex a sentence is.
The above can also be viewed in terms of 'closeness'.
The example illustrates that current syntactic models give no notion of how 
'close' determiners and nouns are.   However ultrametrics do give an 
indication of closeness and this can be compared:
{\sl firstly} to the closeness indicated by features,
{\sl secondly} to the idea that if a sentence is not sufficiently close
then there is a {\sc barrier} Chomsky (1986b) \cite{bi:chomsky2}.
Only the first is looked at here. 
In traditional syntax phrases can be iteratively 
embedded to give sentences of unbounded length and complexity.
A degree of sentence complexity perhaps corresponds to the height of the tree
representing the sentence.   As people can only process a finite amount
of information this height must be finite,
in the traditional theoretical framework
there is no finite bound on sentence length.
An upper bound could perhaps be found by experiment,
inspection of phrase trees suggests
a first guess of $h=12$.\\
The {\sf third} is that it means that syntax is described in the same 
formalism as a lot of other science,  
for example those topics described in the first paragraph,
so that there is the possibility of techniques being used in one area
being deployed in another.\\
The {\sf fourth} is that an ultrametric formulation might allow a 
generalization so that ideas in syntax can be applied to other 
cognitive processes.\\
The {\sf fifth},  and perhaps the most important,
is that it might be possible to use some sort of minimum distance principle
in syntax,  
indeed it could be this minimum description which would have application in
other cognitive processes.\\
\subsection{Ockams Razor.}
Minimum description in science go back several hundred years
to ``{\em Ockams razor}'' or perhaps further,  
see for example Sorton (1947) \cite{sorton} page 552.
The principle of least action
(see for example Bjorken and Drell (1965) \cite{bi:BD} \S 11.2),  
in physics is that minimal variation of a given action gives
field equations which describe the dynamics of a system.  For example,
Maxwell's equations can be derived from a simple action by varying it.
In the present context one would hope that syntax allows for
a minimum encoding of semantic information,
the minimum encoding being given by some ultrametric measure.
A different approach along these lines is that of 
Rissanen (1982) \cite{bi:rissanen} and Zadrozny (2000) \cite{bi:zadrozny}.
Briefly they assign a length of $1$ to each symbol in a sentence,
then the {\sc minimum description length} states that the best theory 
to explain a set of data is the one which minimizes both the sum of:
i) the length,  in bits,  of the description of the theory,  and
ii) the length,  in bits,  of data when encoded with the help of the theory.
Christiansen (2001) \cite{christiansen} discusses how constraint handling
rules (CHR) can be applied to grammars,  this can be thought of as a
minimizing procedure.
\subsection{Recticulate \& N-ary Trees.}
A {\sc reticulate} tree is a tree in which the branches reconverge,  
illustrated by {\bf Figure 2},
\begin{figure}
   {\epsfig{figure=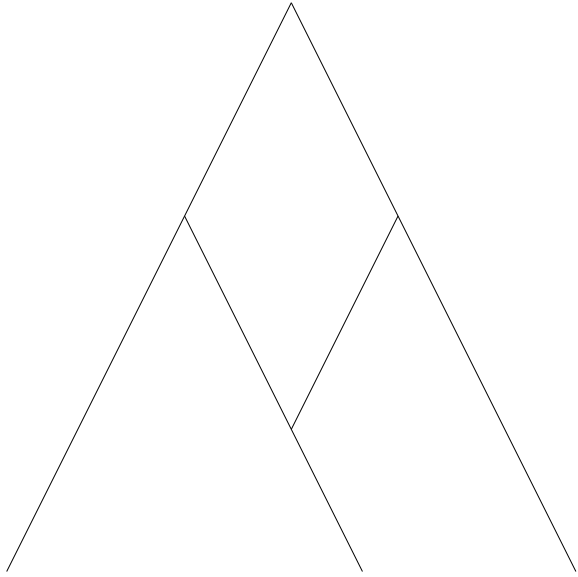,height=3in}}
   \caption{A {\sc reticulate} tree}
\end{figure}
a {\sc non-reticulate} tree is a tree in which the branches do not reconverge. 
{\sc N-ary} branching is illustrated by {\bf Figure 3}. 
\begin{figure}
   {\epsfig{figure=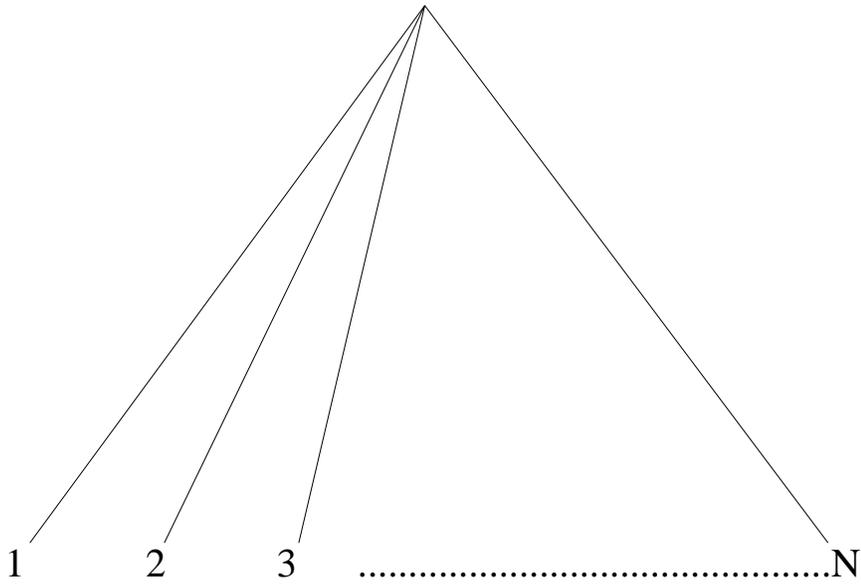,height=3in}}
   \caption{{\sc N-ary} branching}
\end{figure}
{\sc Binary} branching is {\sc N-ary} branching with $N=2$.
{\sc N-ary} branching can be replaced by binary branching 
if additional layers are used.   
A {\sc switched} tree is a tree in which all the branches are binary.
Syntactic phrase trees are {\sc non-reticulate} and {\sc switched}.
In most linguistic theories all syntactic phrase trees have \={X} structure,
Jackendoff (1977) \cite{bi:jackendoff},
here attention is restricted to theory which has \={X} structure.
{\subsection{Sectional Contents.}
In section \ref{sec:sec2} it is shown how to represent trees by matrices 
and triangles.   All \={X} triangles are isosceles but not equilateral.
In section \ref{sec:sec3} the matrix {\bf U} for the minimum ultrametric 
distance for lexical categories is given.
For simplicity discussion is limited 
to active voice sentences with only determiners,  nouns,  transitive verbs,
adjectives,  and prepositions.   
Inclusion of case theory,  COMP,  INFL,.. might be of interest 
but would complicate matters.
In section \ref{sec:sec4} the singular matrix {\bf F} for features is given.
{\bf F} is not an ultrametric matrix and there appears to be no relation 
to {\bf U}.   In section \ref{sec:sec5}
it is shown that the notion of {\sc c-command} is equivalent to 
an ultrametric minimum distance.   This allows a new definition of 
government to be given.   
In appendix \ref{sec:app} other linguistic hierarchies are discussed;  
in particular there appears to be at least two separate 
occurrences of culturally determined partial ordered hierarchies 
- the {\sl accessibility hierarchy} for relative clauses 
and the {\sl universal colour ordering}.    
For completion in appendix \ref{sec:app} there is a very briefly
account what these hierarchies are, a comparison and contrasting of them,  
and the speculation that they are specific examples of a 
{\sl grand cultural hierarchy}.
The question arises of why such hierarchies should exist,
and it might be because they reduce the amount of memory 
needed to process information by clumping information together 
in the style of Miller (1956) \cite{bi:miller},  
for a more recent reference see Cowan (2001) \cite{cowan}.
A {\sl hierarchy} is an example of a representation as
discussed by Roberts (1998) \cite{bi:roberts}.
\section{ \={X} Structure Implies No Equilateral Triangles.}
\label{sec:sec2}
\subsection{Binary and N-ary Branching for simple sentences.}
\={X} structure implies {\sc binary branching}  Haegeman (1994) p.139 
\cite{bi:haegeman}.
To see what this implies for ultrametric 
distances consider all five species of {\sc binary branched} tree,
the {\sl first} has diagram {\bf Figure 4} and corresponding matrix:
\begin{figure}
{\epsfig{figure=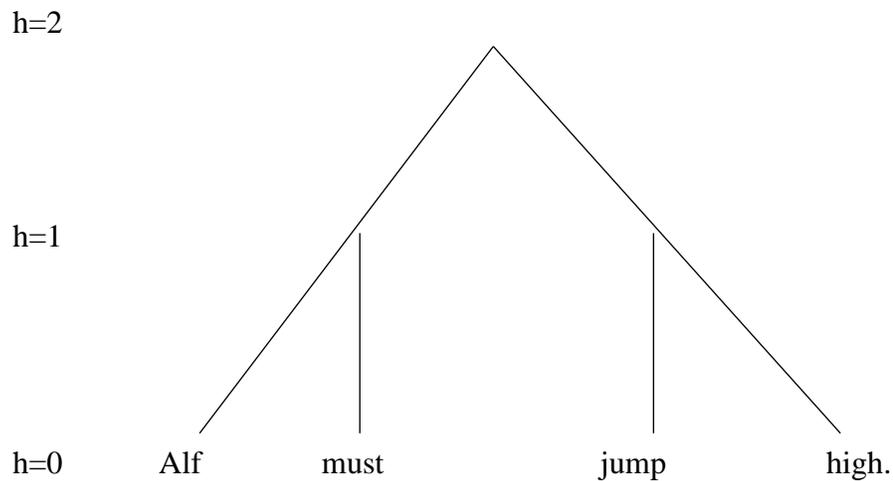,height=2.5in}}
   \caption{The simplest {\sc binary} tree for ``Alf must jump high''}
\end{figure}
\ber
First=
\begin{array}{ccccc}
\bullet & A & M & J & H\\
A & 0 & 1 & 2 & 2\\
M & . & 0 & 2 & 2\\
J & . & . & 0 & 1\\
H & . & . & . & 0\\
\end{array}
\label{eq2}
\ear
respectively.   The matrices corresponding to the other {\sl four}
{\sc binary branched} trees are:
\ber
Second=
\begin{array}{ccccc}
\bullet & A & M & J & H\\
A & 0 & 3 & 3 & 3\\
M & . & 0 & 2 & 2\\
J & . & . & 0 & 1\\
H & . & . & . & 0\\
\end{array}
\label{eq3}
\ear
\ber
Third=
\begin{array}{ccccc}
\bullet & A & M & J & H\\
A & 0 & 3 & 3 & 3\\
M & . & 0 & 1 & 2\\
J & . & . & 0 & 1\\
H & . & . & . & 0\\
\end{array}
\label{eq4}
\ear
\ber
Fourth=
\begin{array}{ccccc}
\bullet & A & M & J & H\\
A & 0 & 2 & 2 & 3\\
M & . & 0 & 1 & 3\\
J & . & . & 0 & 3\\
H & . & . & . & 0\\
\end{array}
\label{eq5}
\ear
\ber
Fifth=
\begin{array}{ccccc}
\bullet & A & M & J & H\\
A & 0 & 1 & 2 & 3\\
M & . & 0 & 2 & 3\\
J & . & . & 0 & 3\\
H & . & . & . & 0\\
\end{array}
\label{eq6}
\ear
There are {\sl two} {\sc 3-ary} trees with matrices:
\ber
Sixth=
\begin{array}{ccccc}
\bullet & A & M & J & H\\
A & 0 & 1 & 1 & 3\\
M & . & 0 & 1 & 2\\
J & . & . & 0 & 2\\
H & . & . & . & 0\\
\end{array}
\label{eq7}
\ear
\ber
Seventh=
\begin{array}{ccccc}
\bullet & A & M & J & H\\
A & 0 & 2 & 2 & 2\\
M & . & 0 & 1 & 1\\
J & . & . & 0 & 1\\
H & . & . & . & 0\\
\end{array}
\label{eq8}
\ear
and finally there is {\sl one} {\sc 4-ary} tree with diagram {\bf Figure 5} 
and matrix:
\begin{figure}
   {\epsfig{figure=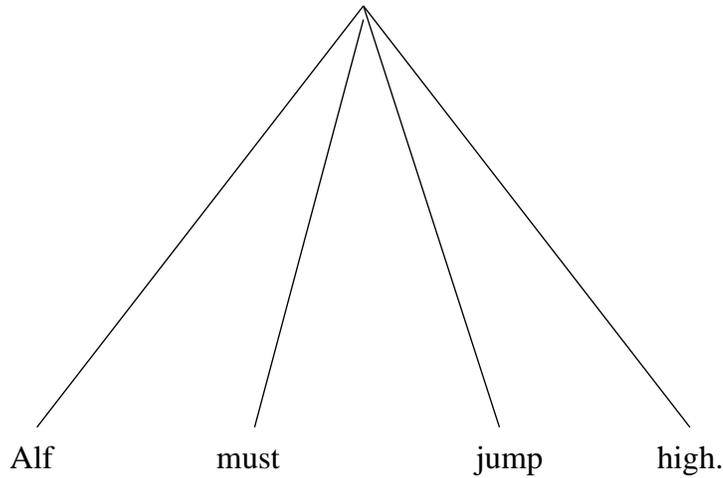,height=2.5in}}
   \caption{The {\sc 4-ary} tree for ``Alf must jump high''}
\end{figure}
\ber
Eighth=
\begin{array}{ccccc}
\bullet & A & M & J & H\\
A & 0 & 1 & 1 & 1\\
M & . & 0 & 1 & 1\\
J & . & . & 0 & 1\\
H & . & . & . & 0\\
\end{array}
\label{eq9}
\ear
\subsection{Triangle representation of the proceeding.}
All ultrametric triangles are isosceles,  but only some are equilateral.
The above suggests that binary branching implies that there are no equilateral 
triangles.   For example from matrix \ref{eq2},  
$d(A,M)=1,  d(A,J)=2,  d(M,J)=2$
giving in the triangle representation {\bf Figure 6},
\begin{figure}
   {\epsfig{figure=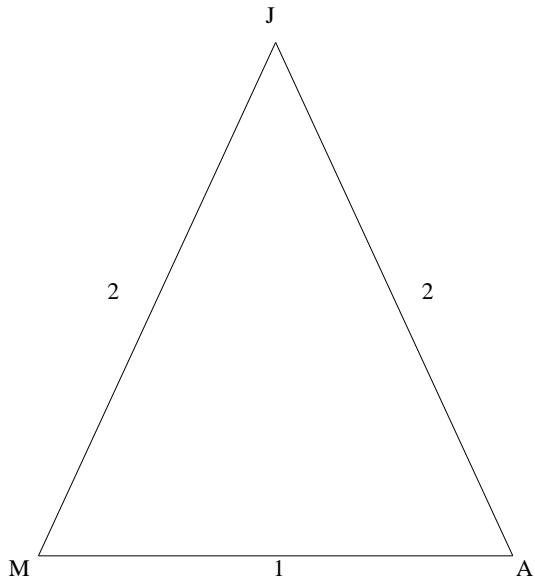,height=3in}}
   \caption{The Isosceles Triangle Representation.}
\end{figure}
and from matrix \ref{eq9},  $d(A,M)=1,  d(A,J)=1,  d(M,J)=1$ 
giving in the triangle representation {\bf Figure 7}.
\begin{figure}
   {\epsfig{figure=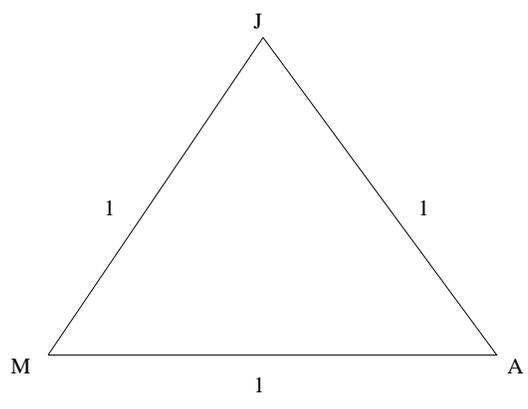,height=2in}}
   \caption{The Equilateral Triangle Representation.}
\end{figure} 
Formally it is proved that \=X structure implies that there are no     
equilateral triangles.
\subsection{The \=X Template.}
Consider the \=X template {\bf Figure 8}.
\begin{figure}
   {\epsfig{figure=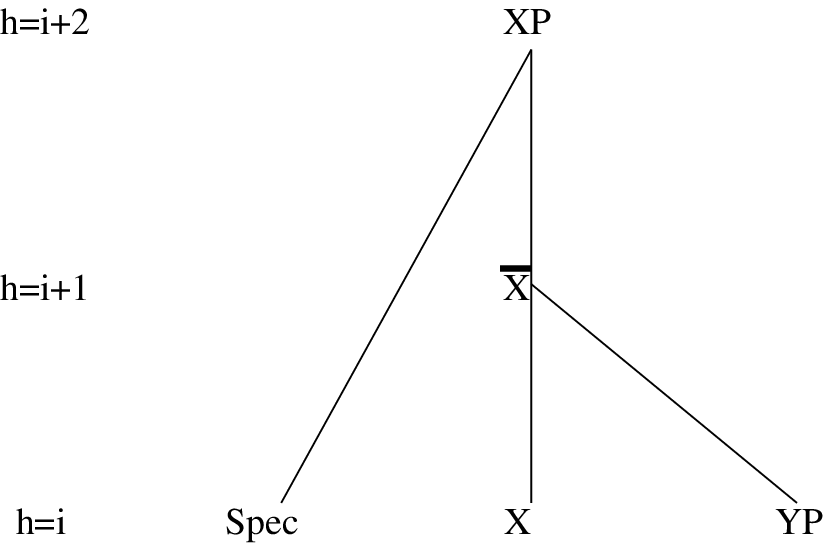,height=3in}} 
   \caption{The \=X Template}
\end{figure} 
\newpage
The matrix representation of this is:
\ber
{\bf \bar{X}}=
\begin{array}{cccc}
\bullet & Spec & X & YP\\
Spec & 0 & i+2 & i+2\\
X & . & 0& i+1\\
YP & . & . & 0\\
\end{array}
\ear
From this the triangle representation is {\bf Figure 9}.
\begin{figure}
   {\epsfig{figure=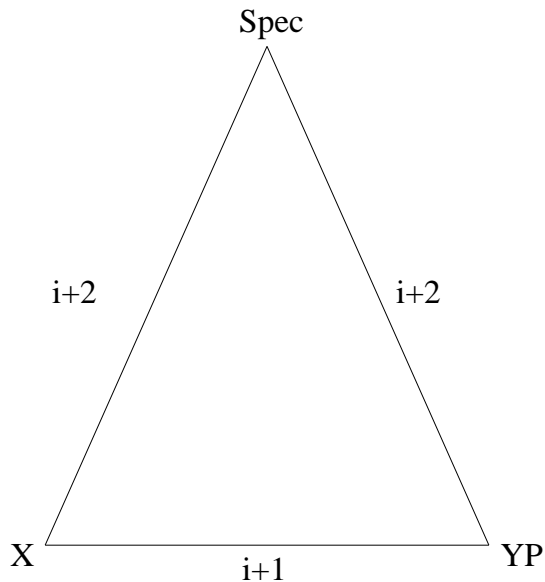,height=3in}} 
   \caption{The Triangle Representation of \=X structure}
\end{figure} 
This must be isosceles but not equilateral.
\section{The minimum ultrametric distance between lexical categories.}
\label{sec:sec3}
\subsection{The minimum distance principle.}
We assume that it is the minimum distance between lexical categories that
is important,  and refer to this as the {\sc minimum distance principle}.
In part this is motivated by the principle of least action in physics,
see for example Bjorken and Drell (1965) \cite{bi:BD} \S 11.2,
see also the introduction \S1.3 above and Roberts (1998) \cite{mdr}\S3.
A current psycholinguistic model of sentence production is the garden path
model,  see for example Frazier (1987) \cite{frazier} 
and  Roberts (1998) \cite{mdr}\S5.4.   Part of this model requires 
{\it the minimal attachment principle},  
which is ``do not postulate unnecessary modes.'':
this can be thought of as a minimum principle.
The {\sc minimum distance principle} implies that the correct tree
for the example in the introduction is {\bf Figure 10},
\begin{figure}
   {\epsfig{figure=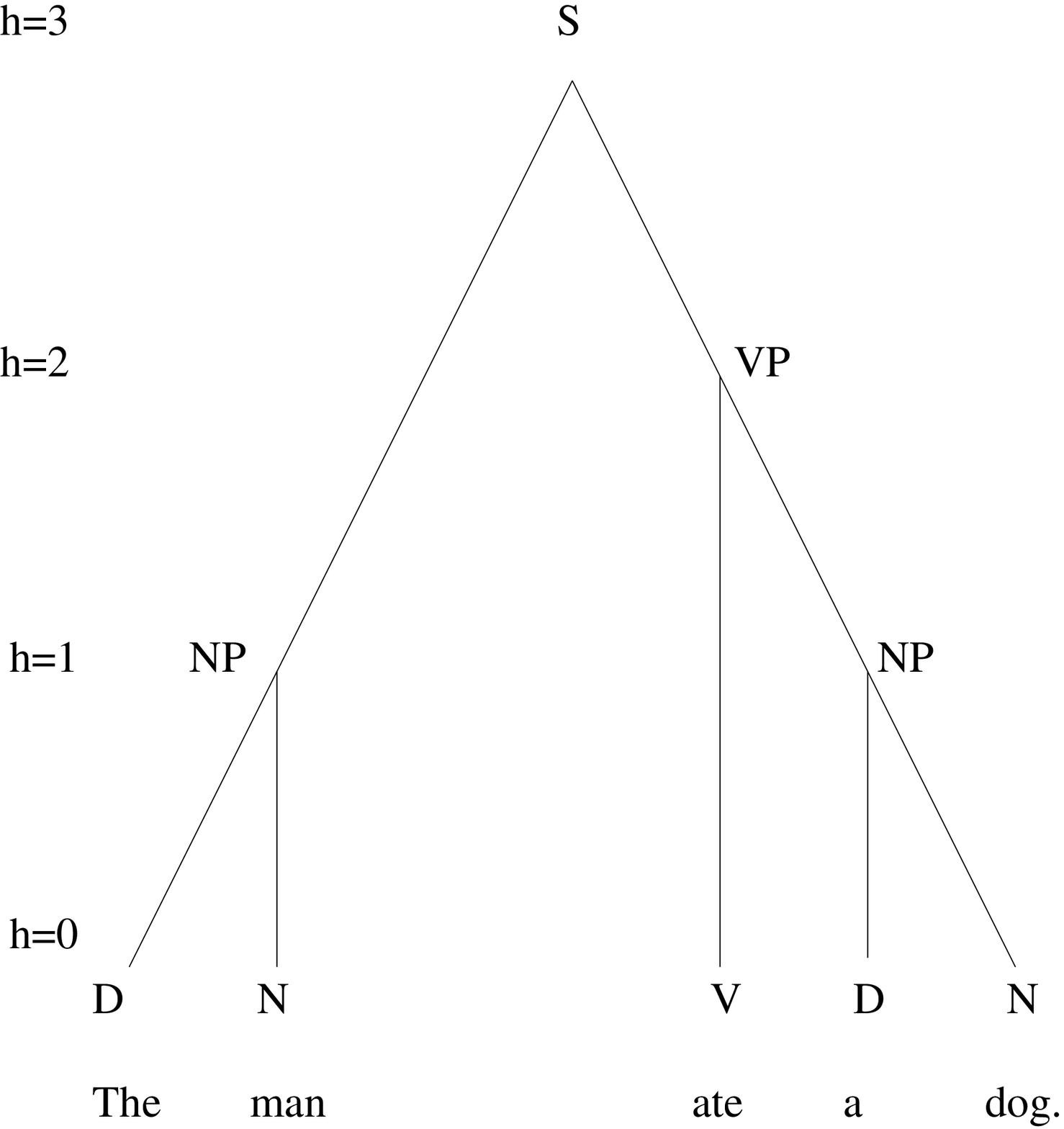,height=4in}}
   \caption{The Correct Tree for the Example in the Introduction.}
\end{figure} 
so that all entries occur at the lowest possible height.
Thus in particular tree {\bf A} is preferred to tree {\bf B}.
This assumption does not effect the matrix {\bf U} given below,
but will have an effect when the analysis is extended to ${\bf\theta}$-theory.
From the above $d(N,D)=1, d(N,V)=d(D,V)=2$.   Similarly from {\bf Figure 11},
\begin{figure}
   {\epsfig{figure=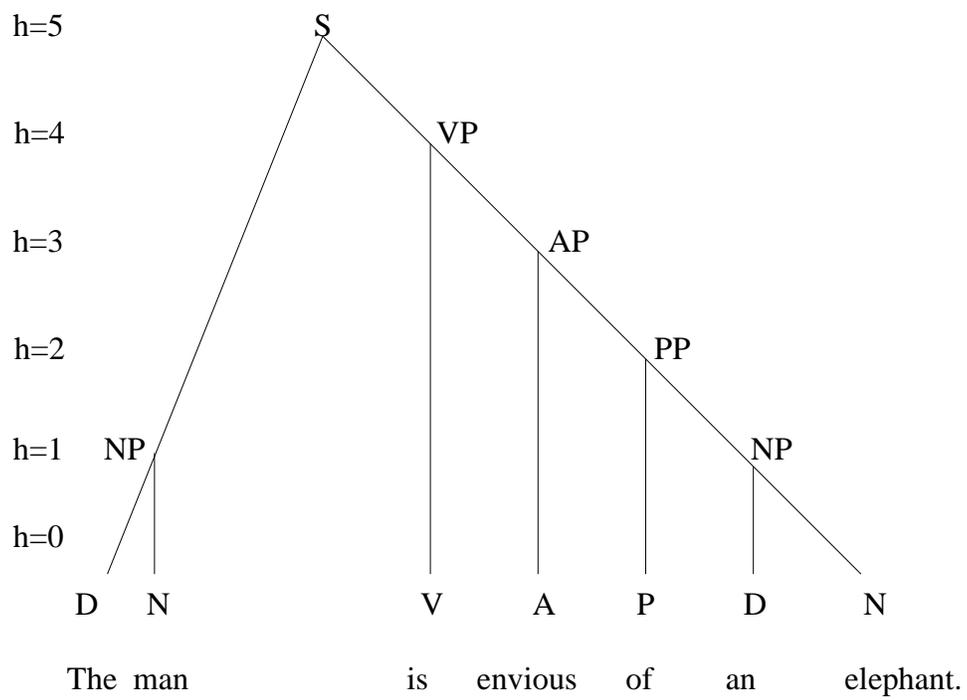,height=3.6in}}
   \caption{The Distance Between Verbs and Adjectives.}
\end{figure} 
$d(V,A)=4$.
Constructing other examples gives the ultrametric distance matrix
\ber
{\bf U}=
\begin{array}{cccccc}
\bullet & D & N & V & A & P\\
D & 0 & 1 & 2 & 2 & 2\\
N & . & 0 & 2 & 2 & 2\\
V & . & . & 0 & 4 & 3\\
A & . & . & . & 0 & 3\\
P & . & . & . & . & 0\\
\end{array}
\label{matrixU}
\ear
Ignoring D and ordering the matrix NPVA (noun,  pronoun,  verb,  adjective)
suggests the pattern
\ber
{\bf I}=
\begin{array}{ccccc}
0 & i & i & i & \dots\\
. & 0 & i+1 & i+1 & \dots\\
. & . & 0   & i+2 & \dots\\
. & . & .   & 0 & \dots\\
\end{array}
\label{matrixI}
\ear
which is compatible with the \={X} matrix of the last section;
however it does not follow by necessity as the \={X} 
case holds for a single sentence and {\bf U} is constructed from several.
\section{Features}\label{sec:sec4}
\subsection{No square matrix representation of Features.}
This section investigates whether there is a general framework
which can describe the preceding and also features.
Haegeman (1994) \cite{bi:haegeman} p.146 
gives the following diagram for features:
\ber
{\rm Features~ diagram} =
\begin{array}{ccc}
Noun & +N & -V\\
Verb & -N & +V\\
Adj. & +N & +V\\
Pre. & -N & -V\\
\end{array}
\ear
This can be represented by the matrix
\ber
{\rm Features~ matrix} =
\begin{array}{ccc}
\cdot    & Noun & Verb\\
Noun & +1   & -1  \\
Verb & -1   & +1  \\
Adj. & -1   & +1  \\
Pre. & -1   & -1  \\
\end{array}
\ear
A square matrix can be constructed by assuming that the matrix is symmetric.
This leaves only one unknown $F(A,P)$.   Taking $F(A,P)= -1$
gives equal number of positive and negative entries in the matrix
\ber
{\bf F} =
\begin{array}{ccccc}
\bullet & N  & V  & A  & P\\
N & +1 & -1 & +1 & -1\\
V & -1 & +1 & +1 & -1\\
A & +1 & +1 & +1 & -1\\
P & -1 & -1 & -1 & +1\\
\end{array}
\label{matrixF}
\ear
which is singular as its determinant vanishes.   There appears to be no 
relation between matrix {\bf F} \ref{matrixF} and matrix {\bf U} \ref{matrixU}.
Using the Pauli matrices 
(see for example Bjorken and Drell (1965) \cite{bi:BD} p.378)
\ber
I=
\begin{array}{cc}
1 & 0\\
0 & 1\\
\end{array}~~~
\sigma^1=
\begin{array}{cc}
0 & 1\\
1 & 0\\
\end{array}~~~
\sigma^2=
\begin{array}{cc}
0 & -i\\
i & 0\\
\end{array}~~~
\sigma^3=
\begin{array}{cc}
1 & 0\\
0 & -1\\
\end{array}
\ear
{\bf F} can be expressed as
\ber
{\bf F}=
\begin{array}{cc}
I-\sigma^1          & -i\sigma^2+\sigma^3\\
+i\sigma^2+\sigma^3 & I-\sigma^1\\
\end{array}
\ear
however this does not correspond in any straightforward way to any of the 
Dirac matrices (see for example Bjorken and Drell (1965) \cite{bi:BD} page 378)
in standard representations.
\section{Ultrametric Approach to Government.}\label{sec:sec5}
Recall the following definitions in Haegeman \cite{bi:haegeman}:
\subsection{Definition of {\sc dominates}.}
{\bf Definition} \cite{bi:haegeman}  p.85  \newline
Node {\bf A} {\sc dominates} node {\bf B} iff:\newline
       i)$~~h(A)$ is higher up or at the same height on the tree as $h(B)$\\
         i.e.$h(A)\geq h(B)$\newline
       ii) it is possible to trace a line from {\bf A} to {\bf B} 
           going only downward,\newline  
         $~~~~$ or at most going to one higher node.\newline
{\bf Remarks}\newline
The {\sl first} requirement is that {\bf A} is at a greater height than
{\bf B}.   The {\sl second} requirement restricts the possible downward route 
from {\bf A} to {\bf B} so that it contains at most one upward segment.
\newline {\bf Example} (compare \cite{bi:haegeman} p.83)\newline
  the phrase tree {\bf Figure 12}
\begin{figure}
   {\epsfig{figure=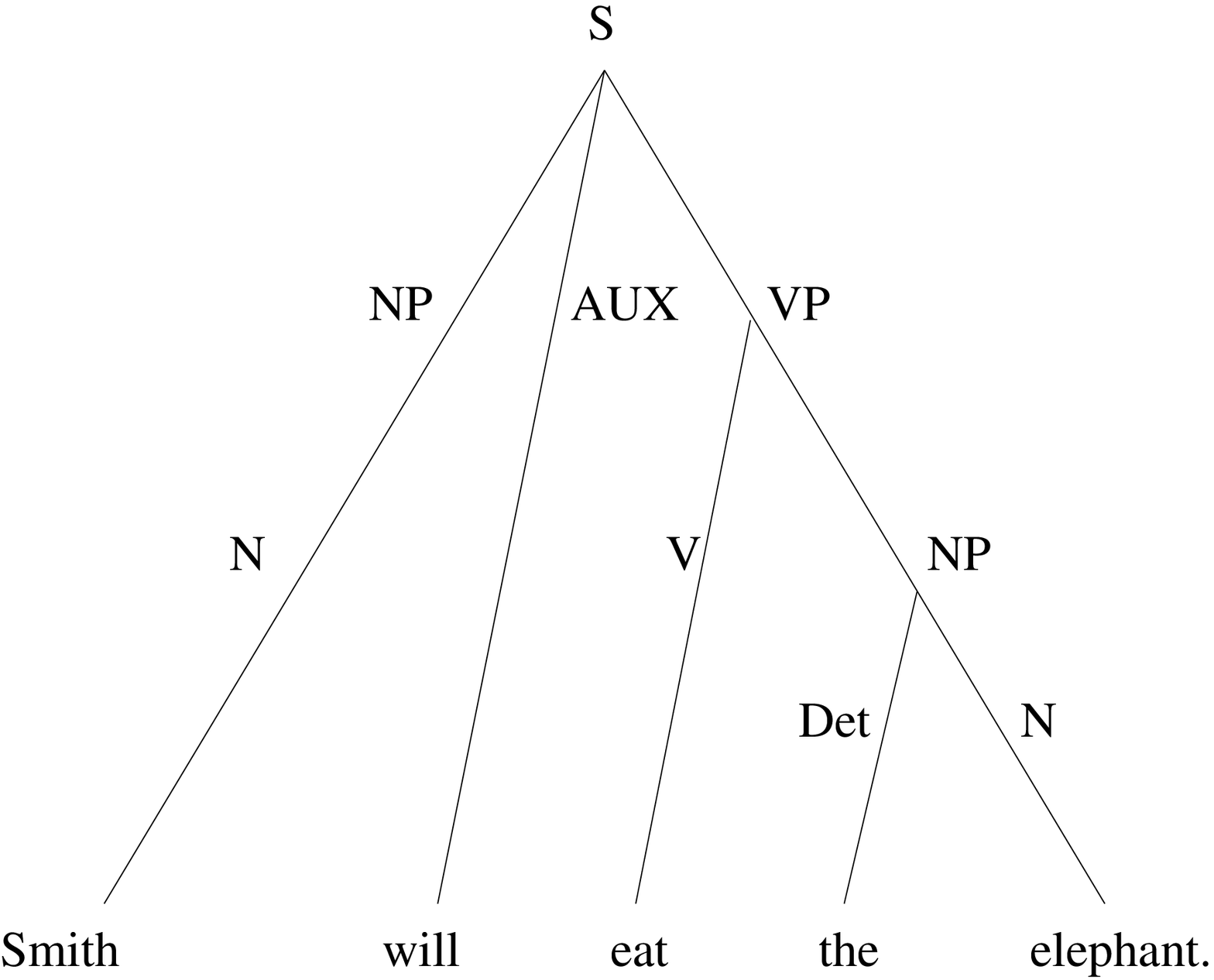,height=3in}}               
   \caption{Illustration of {\sc dominates}}
\end{figure} 
gives the 'dominates' matrix
\ber
{\bf D}=
\begin{array}{cccccccccc}
\bullet & S & NP(S) & N(S) & AUX & VP & V & NP(E) & Det & N(e)\\
S       & 1 & 1     & 1    & 1   & 1  & 1 & 1     & 1   & 1   \\
NP(S)   & 0 & 1     & 1    & 0   & 0  & 0 & 0     & 0   & 0   \\
N(S)    & 0 & 0     & 1    & 0   & 0  & 0 & 0     & 0   & 0   \\
AUX     & 0 & 0     & 0    & 1   & 0  & 0 & 0     & 0   & 0   \\
VP      & 0 & 0     & 0    & 0   & 1  & 1 & 1     & 1   & 1   \\
V       & 0 & 0     & 0    & 0   & 0  & 1 & 1     & 1   & 1   \\
NP(E)   & 0 & 0     & 0    & 0   & 0  & 0 & 1     & 1   & 1   \\
Det     & 0 & 0     & 0    & 0   & 0  & 0 & 0     & 1   & 0   \\
N(E)    & 0 & 0     & 0    & 0   & 0  & 0 & 0     & 0   & 1   \\ 
\end{array}
\ear
where 1 indicates {\bf A} dominates {\bf B} and 0 indicates that it does not.
\subsection{Definition of {\sc C-command}.}
{\bf Definition} \cite{bi:haegeman} p.134 \newline
Node {\bf A} {\sc c-commands} (constituent-commands) node {\bf B} iff:\newline
       i) {\bf A} does not dominate {\bf B} 
           and {\bf B} does not dominate {\bf A},\newline
       ii) The first branching node dominating {\bf A} 
           also dominates {\bf B}.\newline
{\bf Remarks} \newline
The {\sl first} requirement is that there is no direct route up or 
down from {\bf A} to {\bf B}  passing more than one higher node.
The {\sl second} requirement restricts {\bf A} and {\bf B} to be 'close'.   
Haegeman's first criteria for dominance needs to be adjusted,
if it is correct then $h(A)>h(B)$ and $h(B)>h(A)$ so that the set of 
all {\sc c-commands} is empty,  therefore greater than or equal $\geq$ 
is used here instead of greater than $>$.   
Haegeman's second criteria for dominance also needs to be adjusted,
if no higher node is allowed the set of {\sc c-commands} is again empty.   
Chomsky (1986a) \cite{bi:chomsky} p.161 approaches the 
subject in a different manner using maximal projections.\\
{\bf Example}:
{\bf Figure 13}
\begin{figure}
   {\epsfig{figure=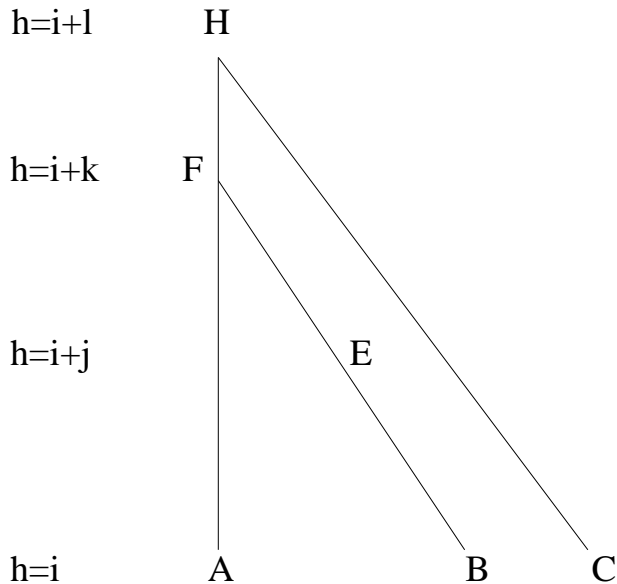,height=3in}}
   \caption{Example of {\sc c-commands}.}
\end{figure} 
in the figure  $0<j<k<l$.   The corresponding ultrametric matrix is
\ber
{\bf U}=
\begin{array}{ccccc}
\bullet & A & B & C & D\\
A       & 0 & k & k & l\\
B       & . & 0 & j & l\\
C       & . & . & 0 & l\\
D       & . & . & . & 0\\
\end{array}
\ear
The {\sc c-command} matrix {\bf CM} is
\ber
{\bf CM}=
\begin{array}{ccccc}
\bullet & A & B & C & D\\
A       & 1 & 1 & 1 & 0\\
B       & 0 & 1 & 1 & 0\\
C       & 0 & 1 & 1 & 0\\
D       & 1 & 1 & 1 & 1\\
\end{array}
\ear
where 1 indicates {\bf A},{\bf B},{\ldots} 
{\sc c-commands} {\bf A},{\bf B},{\ldots}
and 0 indicates that it does not.
\subsection{Definitions of {\sc C-Domain} \& {\sc Governs}.}
{\bf Definition} \cite{bi:haegeman} p.134 \\
The total of all the nodes {\sc c-commanded} by an element is the 
{\sc c-domain} of that element.\\
{\bf Definition} \cite{bi:haegeman} p.135 \\  
{\bf A} {\sc governs} {\bf B} iff:\\
       i) {\bf A} is a {\sc governor},\\
       ii){\bf A} {\sc c-commands} {\bf B} 
      and {\bf B} {\sc c-commands} {\bf A}.\\
{\bf Remarks}:\\
The {\sl first} requirement is a restriction on the set {\bf A} 
(in linguistic terminology the category {\bf A}).
A {\sc governor} is a part of speech which generalizes 
the notion of a verb governing an object;  
unfortunately there does not seem to be a formal definition of it.
The {\sl second} requirement is that {\bf A} and {\bf B} 
should be sufficiently 'close'.
\subsection{Definitions of {\sc CU-Domain} \& {\sc CU-Command}.}
Now let {\bf D(A)} be the set of all the ultrametric distances to other nodes 
at the same height and let {\bf M(A)} be the set of these which have the 
smallest value.\\
Call {\bf M(A)} the {\sc cu-domain} of {\bf A} 
and say {\bf A} {\sc cu-commands} all ${\bf B} \varepsilon {\bf M(A)}$,\\
this is illustrated by {\bf Figure 14}.
\begin{figure}
   {\epsfig{figure=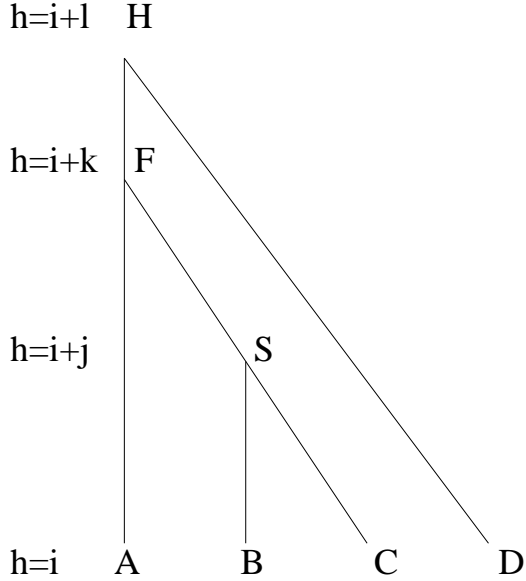,height=3in}}
   \caption{Illustration of the Theorem.}
\end{figure} 
\subsection{Theorem showing the identity between 
{\sc C-Domain} \& {\sc CU-Domain}.}
{\bf Theorem}:\\
The sets {\bf A} {\sc c-commands} {\bf B} 
and {\sc A cu-commands} {\bf B} are identical,
likewise the {\sc c-domain} and the {\sc cu-domain}.\newline
{\bf Proof}:\newline
From the i) part of the definition of {\sc c-command} $h(A)=h(B)$,
so that we are only concerned with nodes at the same height $h(A)=i$.
Let the first branching node above {\bf A} be
{\bf F},  with $h(F)=i+k$.  Let {\bf H} be any node dominating {\bf F},
with $h(H)=i+l$.   Let {\bf E} be the subsidiary node
dominating {\bf B} and {\bf C} and dominated by {\bf F},  with $h(E)=i+j$.
The closest nodes to {\bf A} 
are {\bf B} and {\bf C} both with an ultrametric distance $k$.
The sets {\bf D(A)} and {\bf M(A)} are
{\bf D(A)}=\{{\bf A,B,C,D}\},
{\bf M(A)}=\{{\bf A,B,C}\}.
{\bf A} both {\sc c-commands} and {\sc cu-commands} 
itself and {\bf B} and {\bf C}.
The actual integer values $i,j,k,\ldots$ are arbitrary 
thus the result holds in general.
\subsection{A New Definition of Government.}
This allows a new definition of {\sc government}.
{\bf A} {\sc governs} {\bf B} iff:\\
i) {\bf A} is a {\sc governor}.\\
ii) both ${\bf A} \varepsilon {\bf M(B)}$ 
     and ${\bf B} \varepsilon {\bf M(A)}$.\\
This definition of {\sc government} is the same as 
the previous definition of {\sc government}, 
but with the {\sc c-command} requirement replaced by an ultrametric 
requirement that distances be minimal.
\section{Appendix:  Other Linguistic Hierarchies}
\label{sec:app}
\subsection{The Accessibility Hierarchy.}
A {\sc relative clause} ({\em RC}) 
is a clause that modifies a noun or pronoun that
occurs elsewhere in a sentence.   
The {\sl accessibility hierarchy} ({\em AH}) for relative clauses is given by
Keenan and Comrie (1977) \cite{bi:KC} and illustrated in {\bf Figure 15}.
\begin{figure}
   {\epsfig{figure=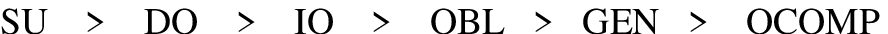,height=0.17in}}
   \caption{The {\sl accessibility hierarchy}}
\end{figure} 
Noun phrases (NP) occurring to the left of "$>$" are more accessible 
than those on the right.   SU is  short for subject,  
DO for direct object,   IO for indirect object,   
OBL for major oblique case NP,  GEN for genitive NP, 
OCOMP for object of comparison.   
The properties of the accessible hierarchy 
are contained in two sets of constraints.\\   
The accessible hierarchy constraints ({\sl AHC's}) are:\\
{\bf AHC1)} A language must be able to relativize subjects.\\
{\bf ACH2)} Any RC forming strategy must apply to a 
continuous segment of the AH.\\
{\bf ACH3)}  Strategies that apply at one point of the 
AH may in principle cease to apply  
at any lower point.\\
The primary relativization constraints ({\sl PRC's}) are\\
{\bf PRC1}) A language must have a primary RC-forming strategy.\\
{\bf PRC2)} If a primary strategy in a given language  
can apply to a low position on the AH,  then it 
can apply to all higher positions.\\
{\bf PRC3)}  A primary strategy may cut off at any point on the AH.\newline
For a given language a deployment that can be used to relativize a clause 
at a specified place on the AH can also be used to relativize all more
accessible clauses.   The type of relativization varies from language to 
language.   There appears to be nothing known on how the skill 
to deploy a relativization develops in an individual.   
One would expect that when a given method is applied the
less accessible would take longer to process, there seems to be no 
psycholinguistic tests done to see if this is indeed the case.
\subsection{The Berlin-Kay Universal Colour Partial Ordering.}
The perception of colour often involves the deployment of a colour name
strategy.   The effect of this is to alter the way the colour is perceived.
The five principles of colour perception are:\\
{\bf CP1)}The communicability of a referent in an array and for a particular
community is 
very closely related to the memorability of that referent in
the same array and for members of the same community.\\
{\bf CP2)}  In the total domain of colour there are 
eleven small focal areas in which are found 
the best instances of the colour categories named in 
any particular language.   
The focal areas are human universals,  
but languages differ in the number of  
basic colour terms they have:  
they vary from two to eleven.\\
{\bf CP3)}  Colour terms appear to evolve in a language 
according to the Berlin-Kay (1969)  
\cite{bi:BK}
universal partial ordering illustrated by {\bf Figure 16},
\begin{figure}
   {\epsfig{figure=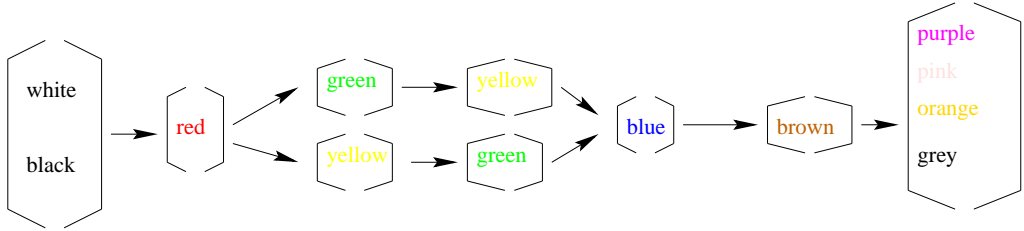,height=1.2in}}
   \caption{The Berlin-Kay Universal Colour Partial Ordering}
\end{figure} 
{\bf CP4)}  Focal colours are more memorable and easier 
to recognize than any other colours,  
whether or not the subject speak a 
language having a name for the colour.\\
{\bf CP5)}  The structure of the colour space determined by
multi-dimensional scaling of  
perceptual data is probably the 
same for all human communities and it is  
unrelated to the space yielded by naming data.\\
Again there is a {\sl culturally determined linguistic partial ordering} (or
hierarchy).   On this occasion it determines the semantic content of 
individual words rather than syntax rules.   Again there appears to be
nothing known on how the skill develops in an individual,  or any timing
tests on the possession of a colour name strategy.   The existence of two 
separate hierarchical {\sl partial orderings} suggests that there is a general 
mechanism for there construction.   Most members of a community seem to develop
these culturally determined skills suggesting that the capacity to develop them
is usually innate but their manifestation depends on environment.

\end{document}